# A Theory of Goal-Oriented MDPs with Dead Ends

**Andrey Kolobov**  **Mausam**  **Daniel S. Weld**
{akolobov, mausam, weld}@cs.washington.edu
Dept of Computer Science and Engineering
University of Washington
Seattle, USA, WA-98195

## Abstract

Stochastic Shortest Path (SSP) MDPs is a problem class widely studied in AI, especially in probabilistic planning. They describe a wide range of scenarios but make the restrictive assumption that the goal is reachable from any state, i.e., that dead-end states do not exist.Because of this, SSPs are unable to model various scenarios that may have catastrophic events (e.g., an airplane possibly crashing if it flies into a storm). Even though MDP algorithms have been used for solving problems with dead ends, a principled theory of SSP extensions that would allow dead ends, including theoretically sound algorithms for solving such MDPs, has been lacking. In this paper, we propose three new MDP classes that admit dead ends under increasingly weaker assumptions. We present Value Iteration-based as well as the more efficient heuristic search algorithms for optimally solving each class, and explore theoretical relationships between these classes. We also conduct a preliminary empirical study comparing the performance of our algorithms on different MDP classes, especially on scenarios with unavoidable dead ends.

## 1 Introduction

Stochastic Shortest Path (SSP) MDPs [Bertsekas, 1995] is a class of probabilistic planning problems thoroughly studied in AI. They describe a wide range of scenarios where the objective of the agent is to reach a goal state in the least costly way in expectation from any non-goal state using actions with probabilistic outcomes.

While SSPs are a popular model, they have a serious limitation. They assume that a given MDP has at least one *complete proper policy*, a policy that reaches the goal from any state with 100% probability. Basic algorithms for solving SSP MDPs, such as Value Iteration (VI) [Bellman, 1957], fail to converge if this assumption does not hold. In the meantime, this requirement effectively disallows the existence of *dead ends*, states from which reaching the goal is impossible, and of catastrophic events that lead to these states. Such catastrophic failures are a possiblity to be reckoned with in many real-world planning problems, be it sending a rover on Mars or navigating a robot in a building with staircases. Thus, insisting on the absence of dead ends significantly limits the applicability of SSPs. Moreover, verifying that a given MDP has no dead ends can be nontrivial, further complicating the use of this model.

Researchers have realized that allowing dead ends in goal-oriented MDPs would break some existing methods for solving them [Little and Thiebaux, 2007]. They have also suggested algorithms that are aware of the possible presence of dead-end states [Kolobov et al., 2010] and try to avoid them when computing a policy [Keyder and Geffner, 2008, Bonet and Geffner, 2005]. However, these attempts have lacked a theoretical analysis of how to incorporate dead ends into SSPs in a principled way, and of what the optimization criteria in the presence of dead ends should be. This paper bridges the gap by introducing three new MDP classes with progressively weaker assumptions about the existence of dead ends, analyzing their properties, and presenting optimal VI-like and the more efficient heuristic search algorithms for them.

The first class we present, SSPADE, is a small extension of SSP that has well-defined easily-computable optimal solutions if dead ends are present but are avoidable *provided that the process starts at a known initial state $s_0$*.

The second and third classes introduced in this paper admit that dead ends may exist and the probability of running into them from the initial state may be positive no matter how hard the agent tries. If the chance of a catastrophic event under any policy is nonzero, a key question is: should we prefer policies that minimize the expected cost of getting to the goal even at the expense of an increased risk of failure, or those that reduce the risk of failure above all else?

The former criterion characterizes scenarios where entering a dead end, while highly undesirable, has a finite "price". For instance, suppose the agent buys an expensive ticket for a concert of a favorite band in another city, but remem-bers about it only on the day of the event. Getting to the concert venue requires a flight, either by hiring a business jet or by a regular airline with a layover. The first option is very expensive but almost guarantees making the concert on time. The second is much cheaper but, since the concert is so soon, missing the connection, a somewhat probable outcome, means missing the concert. Nonetheless, the cost of missing the concert is only the price of the ticket, so a rational agent would choose to travel with a regular airline. Accordingly, one of the MDP classes we propose, fSSPUDE, assumes that the agent can put a price (penalty) on ending up in a dead end state and wants to compute a policy with the least expected cost (including the possible penalty). While seemingly straightforward, this intuition is tricky to operationalize because of several subtleties. We overcome these subtleties and show how fSSPUDE can be solved with easy modifications to existing SSP algorithms.

In the third MDP class we introduce, iSSPUDE, not only are dead ends unavoidable, but the cost of hitting one is assumed to be infinitely large. Consider, for example, the task of planning an ascent to the top of Mount Everest for a group of human alpinists. Such an ascent is fraught with inherent lethal risks, and to any human, the price of their own life can be taken as infinite. Note the conceptual difficulty with this setting: since every policy reaches an infinite-cost state, the expected cost of any policy is also infinite. This makes SSP's cost-minimization criterion uninformative. Instead, for an undertaking as above, a natural primary objective is to maximize the *probability* of getting to the goal (i.e., to minimize the chance of getting into a lethal accident, a dead-end state). However, of all policies maximizing this chance, we would prefer the least costly one (in expectation). This is exactly the multiobjective criterion we propose for this class of MDPs. Solving iSSPUDE is much more involved than handling the previous two classes, and we introduce two novel algorithms for it.

Intuitively, the objectives of fSSPUDE and iSSPUDE MDPs are related — as fSSPUDE's dead-end penalty gets bigger, the optimal policies of the two classes coincide. We provide a theoretical and an empirical analysis of this insight, showing that solving fSSPUDE yields an optimal policy for iSSPUDE if the dead-end penalty is high enough.

Thus, the paper makes four contributions: (1) three new goal-oriented MDP models that admit the existence of dead-end states; (2) optimal VI and heuristic search algorithms for solving them; (3) theoretical results describing equivalences among problems in these classes; and (4) an empirical evaluation tentatively answering the question: which class should be used when modeling a given scenario involving unavoidable dead ends?

## 2 Background and Preliminaries

**SSP MDPs.** In this paper, we extend an MDP class known as the Stochastic Shortest Path (SSP) problems with an optional initial state, defined as tuples of the form $\langle \mathcal{S}, \mathcal{A}, \mathcal{T}, \mathcal{C}, \mathcal{G}, s_0 \rangle$, where $\mathcal{S}$ is a finite set of states, $\mathcal{A}$ is a finite set of actions, $\mathcal{T}$ is a transition function $\mathcal{S} \times \mathcal{A} \times \mathcal{S} \rightarrow [0, 1]$ that gives the probability of moving from $s_i$ to $s_j$ by executing $a$, $\mathcal{C}$ is a map $\mathcal{S} \times \mathcal{A} \rightarrow \mathbb{R}$ that specifies action costs, $\mathcal{G}$ is a set of (absorbing) goal states, and $s_0$ is an optional start state. For each $g \in G$, $\mathcal{T}(g, a, g) = 1$ and $\mathcal{C}(g, a) = 0$ for all $a \in A$, which forces the agent to stay in $g$ forever while accumulating no reward.

An SSP must also satisfy two conditions: (1) It must have at least one *complete proper policy*, a rule prescribing an action for every state with which an agent can reach a goal state from any state with probability 1. (2) Every *improper* policy must incur the cost of $\infty$ from all states from which it cannot reach the goal with probability 1.

When the initial state is unknown, solving an SSP MDP means finding a policy whose execution from any state allows an agent to reach a goal state while incurring the least expected cost. We call such a policy *complete optimal*, and denote any complete policy as $\pi$. When the initial state is given, we are interested in computing an *optimal (partial) policy rooted at* $s_0$, i.e., one that reaches the goal in the least costly way from $s_0$ and is defined for every state it can reach from $s_0$ (though not necessarily for other states).

To make the notion of policy cost more concrete, we define a *cost function* as a mapping $J : \mathcal{S} \rightarrow \mathbb{R} \cup \{\infty\}$ and let random variables $S_t$ and $A_t$ denote, respectively, the state of the process after $t$ time steps and the action selected in that state. Then, the cost function $J^\pi$ of policy $\pi$ is

$$J^\pi(s) = \mathbb{E}_s^\pi \left[ \sum_{t=0}^{\infty} \mathcal{C}(S_t, A_t) \right] \quad (1)$$

In other words, the cost of a policy $\pi$ at a state $s$ is the expectation of the total cost the policy incurs if the execution of $\pi$ is started in $s$. In turn, every cost function $J$ has a policy $\pi^J$ that is $J$-greedy, i.e., that satisfies

$$\pi^J(s) = \arg\min_{a \in \mathcal{A}} \left[ \mathcal{C}(s, a) + \sum_{s' \in \mathcal{S}} \mathcal{T}(s, a, s') J(s') \right] \quad (2)$$

Optimally solving an SSP MDP means finding a policy that minimizes $J^\pi$. Such policies are denoted $\pi^*$, and their cost function $J^* = J^{\pi^*}$, called the *optimal cost function*, is defined as $J^* = \min_\pi J^\pi$. $J^*$ also satisfies the following condition, the *Bellman equation*, for all $s \in \mathcal{S}$:

$$J(s) = \min_{a \in \mathcal{A}} \left[ \mathcal{C}(s, a) + \sum_{s' \in \mathcal{S}} \mathcal{T}(s, a, s') J(s') \right] \quad (3)$$

**Value Iteration for SSP MDPs.** The Bellman equation suggests a dynamic programing method of solving SSPs, known as Value Iteration (VI$_{SSP}$) [Bellman, 1957]. VI$_{SSP}$ starts by initializing state costs with an arbitrary *heuristic cost function* $\hat{J}$. Afterwards, it executes several sweeps of the state space and updates every state during every sweep by using the Bellman equation (3) as an assignment operator, the *Bellman backup operator*. Denoting the cost func-tion after the $i$-th sweep as $J_i$, it can be shown that the sequence $\{J_i\}_{i=1}^\infty$ converges to $J^*$. A complete optimal policy $\pi^*$ can be derived from $J^*$ via Equation 2.

**Heuristic Search for SSP MDPs.** Because it stores and updates the cost function for the entire $\mathcal{S}$, VI$_{SSP}$ can be slow and memory-inefficient even on relatively small SSPs. However, if the initial state $s_0$ is given we are interested in computing $\pi^*_{s_0}$, an optimal policy from $s_0$ only, which typically does not visit (and hence does not need to be defined for) all states. This can be done with a family of algorithms based on VI called *heuristic search*. Like VI, these algorithms need to be initialized with a heuristic $\hat{J}$. However, if $\hat{J}$ is *admissible*, i.e., satisfies $\hat{J}(s) \leq J^*(s)$ for all states, then heuristic search algorithms can often compute $J^*$ for the states relevant to reaching the goal from $s_0$ without updating or even memoizing costs for many of the other states. At an abstract level, the operation of any heuristic search algorithm is represented by the FIND-AND-REVISE framework [Bonet and Geffner, 2003a]. As formalized by FIND-AND-REVISE, any heuristic search algorithm starts with an admissible $\hat{J}$ and explicitly or implicitly maintains the graph of a policy greedy w.r.t. the current $J$, updating the costs of states *only in this graph* via Bellman backups. Since the initial $\hat{J}$ makes many states look "bad" a-priori, they never end up in the greedy graph and hence never have to be stored or updated. This makes heuristic search algorithms, e.g., LRTDP [Bonet and Geffner, 2003b], work more efficiently than VI and still produce an optimal $\pi^*_{s_0}$.

**GSSP and MAXPROB MDPs.** Unfortunately, many interesting probabilistic planning scenarios fall outside of the SSP MDP class. One example is MAXPROB MDPs [Kolobov et al., 2011], goal-oriented problems where the objective is to maximize the probability of getting to the goal, not minimize the cost. To discuss MAXPROB, we need the following definition:

**Definition** For an MDP with a set of goal states $\mathcal{G} \subset \mathcal{S}$, the *goal-probability function* of a policy $\pi$, denoted $P^\pi$, gives the probability of reaching the goal from any state $s$. Mathematically, letting $S_t^{\pi_s}$ be a random variable denoting a state the MDP may end up if policy $\pi$ is executed starting in state $s$ for $t$ time steps,

$$P^\pi(s) = \sum_{t=1}^\infty P[S_t^{\pi_s} = g \in \mathcal{G}, S_{t'}^{\pi_s} = s \notin \mathcal{G}\ \forall\ 1 \leq t' < t] \quad (4)$$

Each term in the above summation denotes the probability that, if $\pi$ is executed starting at $s$, the MDP ends up in a goal state at step $t$ and not earlier. Once the system enters a goal state, it stays in that goal state forever, so the sum of all such terms is the probability of the system ever entering a goal state under $\pi$.

To introduce MDPs with dead ends, we will only need the following informal definition of MAXPROB MDPs. A MAXPROB MDP is a problem that can be derived from any goal-oriented MDP (e.g., an SSP) by disregarding the cost function $\mathcal{C}$ and maximizing the *probability* of reaching the goal instead of the expected cost. More specifically, solving a MAXPROB means finding the optimal goal-probability function from the above definition, one that satisfies $P^*(s) = \arg\max_\pi P^\pi(s)$ for all states. Alternatively, $1 - P^*(s)$ can be interpreted as the smallest probability of running into a dead end from $s$ for any policy. Thus, solving a MAXPROB derived from a goal-oriented MDP by discarding action costs can be viewed as a way to identify *dead ends*:

**Definition** For a goal-oriented MDP, a *dead-end state* (or dead end, for short) is a state $s$ for which $P^*(s) = 0$.

MAXPROBs, SSPs themselves, and many other MDPs belong to the broader class of Generalized SSP MDPs (GSSPs) [Kolobov et al., 2011]. GSSPs are defined as tuples $\langle \mathcal{S}, \mathcal{A}, \mathcal{T}, \mathcal{C}, \mathcal{G}, s_0 \rangle$ of the same form as SSPs, but relax both of the additional conditions in the SSP definition. In particular, they do not require the existence of a complete proper policy as SSPs do. To state the main results of this paper, we will not need the specifics of GSSP's technical definition, and will only refer to the GSSP properties and algorithms described below.

**Value Iteration for GSSP MDPs.** In the case of SSPs, VI$_{SSP}$ yields a complete optimal policy for these MDPs independently of the initializing heuristic $\hat{J}$. For a GSSP MDP, such a policy need not exist, so there is no analog of VI$_{SSP}$ that works for all problems in this class. However, for MAXPROB, a subclass of GSSP particularly important to us in this paper, such an algorithm, called VI$_{MP}$, can be designed. Like VI$_{SSP}$, VI$_{MP}$ can be initialized with an arbitrary heuristic function, but instead of the Bellman backup operator it uses its generalized version that we call *Bellman backup with Escaping Traps* (BET) in this paper. BET works by updating the initial heuristic function with Bellman backup, until it arrives at a fixed-point function $P^\times$. For SSPs, Bellman backup has only one fixed point, the optimal $P^*$, so if we were working with SSPs, we would stop here. However, for GSSPs (and MAXPROB in particular) this is not the case — $P^*$ is only *one of* Bellman backup's fixed points, and the current fixed point $P^\times$ may not be equal to $P^*$. Crucially, to check whether $P^\times$ is optimal, BET applies the *trap elimination* operator to it, which involves constructing the transition graph that uses actions of *all* policies greedy w.r.t. $P^\times$. If $P^\times \neq P^*$, trap elimination generates a new, non-fixed-point $P^{\times'}$, on which BET again acts with Bellman backup, and so on. The fact that VI$_{MP}$ and FRET, the heuristic search framework for GSSPs considered below, sometimes need to build a greedy transition graph w.r.t. a cost function is important for analyzing the performance of algorithms introduced in this paper (Section 8).

VI$_{MP}$'s main property, whose proof is a straightforward extension of the results in the GSSP paper [Kolobov et al., 2011], is similar to VI$_{SSP}$'s:

**Theorem 1.** *On MAXPROB MDPs, $VI_{MP}$ converges to the optimal goal-probability function $P^*$ independently of the initializing heuristic function $\hat{J}$.*

**Heuristic Search for GSSP MDPs.** Although a complete optimal policy does not necessarily exist for a GSSP, one rooted at $s_0$ always does and can be found by any heuristic search algorithm conforming to an FIND-AND-REVISE analogue for GSSPs, FRET [Kolobov et al., 2011]. Like FIND-AND-REVISE on SSPs, FRET guarantees convergence to $\pi^*_{s_0}$ if the initializing heuristic is admissible.

## 3 MDPs with Avoidable Dead Ends

All definitions of the SSP class in the literature [Bertsekas, 1995, Bonet and Geffner, 2003a, Kolobov et al., 2011] require that the goal be reachable with 100%-probability from every state in the state space, even when initial state $s_0$ is known and the objective is to find an optimal policy rooted only at that state. We first extend SSPs to the easiest case — when dead ends exist but can be avoided entirely from $s_0$.

**Definition** A Stochastic Shortest Path MDP with Avoidable Dead Ends (SSPADE) is a tuple $\langle \mathcal{S}, \mathcal{A}, \mathcal{T}, \mathcal{C}, \mathcal{G}, s_0 \rangle$ where $\mathcal{S}, \mathcal{A}, \mathcal{T}, \mathcal{C}, \mathcal{G}$, and $s_0$ are as in the SSP MDP definition, under the following conditions:

- The initial state $s_0$ is known.
- There exists at least one proper policy rooted at $s_0$.
- Every *improper* policy has $J^\pi(s_0) = \infty$.

Solving a SSPADE MDP means finding a policy $\pi^*_{s_0}$ rooted at $s_0$ that satisfies $\pi^*(s_0) = \arg\min_\pi J^\pi(s_0)$.

SSPADE has only two notable differences from SSP — the former assumes the initial state to be known, and only requires the existence of a partial proper policy. However, even such small departures from the SSP definition prevent some SSP algorithms from working on SSPADE, as we are about to see.

**Value Iteration:** Even though dead ends may be avoided from $s_0$ with an optimal policy, they are still present in the state space. Thus, $VI_{SSP}$, which operates on the entire state space, does not converge on SSPADEs, since the optimal costs for dead ends are infinite. One might think that we may be able to adapt $VI_{SSP}$ to SSPADE by restricting computation to the subset of states reachable from $s_0$. However, even this is not true, because SSPADE requirements do not preclude dead ends reachable from $s_0$. Overall, for $VI_{SSP}$ to work on SSPADEs, we need to detect divergence of state cost sequences – an unsolved problem, to our knowledge.

**Heuristic Search:** Although VI does not terminate for SSPADE, heuristic search algorithms do. This is because:

**Theorem 2.** $SSPADE \subset GSSP$.

*Proof sketch.* SSPADE directly satisfies all requirements of the GSSP definition [Kolobov et al., 2011]. □

In fact, we can also show that heuristic search for SSPADE only needs the regular Bellman backup operator (instead of the BET operator). That is, the FIND-AND-REVISE framework applies to SSPADE without modification. In particular, FIND-AND-REVISE starts with admissible state costs, i.e., underestimates of their optimal costs. As FIND-AND-REVISE updates them, costs of dead ends grow without bound, while costs of other states converge to finite values. Thus, dead ends become unattractive and drop out of the greedy policy graph rooted at $s_0$, leaving FIND-AND-REVISE with an optimal, proper policy.

At the same time, some of the specific heuristic search algorithms for SSP that implement the FIND-AND-REVISE template may fail to terminate on SSPADE. For instance, LAO* and LRTDP may try to update values of dead ends until convergence, which can never be reached. However, each of them can be easily amended to halt on SSPADE problems with an optimal solution. Thus, the existing heuristic search techniques carry over to SSPADE with little adaptation.

## 4 MDPs with Unavoidable Dead Ends

In this section, our objective is threefold: (1) to motivate the semantics of SSP extensions that admit unavoidable dead ends; (2) to state intuitive policy evaluation criteria and thereby induce the notion of optimal policy for models in which the agent pays finite and infinite penalty for visiting dead ends; (3) to formally define MDP class SSPUDE with subclasses fSSPUDE and iSSPUDE that model such finite- and infinite-penalty scenarios.

As a motivation, consider an *improper* SSP MDP, one that conforms to the SSP definition except for the requirement of proper policy existence, and hence has unavoidable dead ends. In such an MDP, the objective of finding a policy that minimizes the expected cost of reaching the goal becomes ill-defined. It implicitly assumes that for at least one policy, the cost incurred by all of the policy's trajectories is finite; however, this cost is finite only for proper policies, all of whose trajectories terminate at the goal. Thus, all policies in an improper SSP may have an infinite expected cost, making the cost criterion unhelpful for selecting the "best" policy.

We suggest two ways of amending the optimization criterion to account for unavoidable dead ends. The first is to assign a finite positive penalty $D$ for visiting a dead end. The semantics of an improper SSP altered this way would be that the agent pays $D$ when encountering a dead end, and the process stops. However, this straightforward modification to the MDP cannot be directly operationalized, since the set of dead-ends is not known a-priori and needs to be inferred while planning. Moreover, this definition also has a caveat – it may cause non-dead-end states that lie on potential paths to a dead end to have higher costs than dead ends themselves. For instance, imagine a state $s$ whose only action leads with probability $(1 - \epsilon)$ to a dead end, with probability $\epsilon > 0$ to the goal, and costs $\epsilon(D+1)$. A simple calculation shows that $J^*(s) = D + \epsilon > D$, even though reaching the goal from $s$ *is* possible. Moreover, notice that this semantic paradox cannot be resolved just by increasing the penalty $D$, because the cost of $s$ will always exceed the dead-end penalty by $\epsilon$.

Therefore, we change the semantics of the finite-penalty model as follows. Whenever the agent reaches *any* state with the expected cost of reaching the goal equaling $D$ or greater, the agent simply pays the penalty $D$ and "gives up", i.e., the process stops. Intuitively, this setting describes scenarios where the agent can put a price on how desirable reaching the goal is. For instance, in the example from the introduction involving a concert in another city, paying the penalty corresponds to deciding not to go to the concert, i.e., foregoing the pleasure the agent would have derived from attending the performance.

The benefit of putting a "cap" on any state's cost as described above is that the cost of a state under any policy becomes finite, formally defined as

$$J_F^\pi(s) = \min_\pi \left\{ D, \mathbb{E}\left[\sum_{t=0}^{\infty} \mathcal{C}(S_t^{\pi_s}, A_t^{\pi_s})\right]\right\} \quad (5)$$

It can be shown that for an improper SSP, there exists an optimal policy $\pi^{*\,1}$, one that satisfies

$$\pi^*(s) = \arg\min_\pi J_F^\pi(s) \;\forall\; s \in \mathcal{S} \quad (6)$$

As we show shortly, we can find such a policy using the expected-cost analysis similar to that for ordinary SSP MDPs. The intuitions just described motivate the fSSPUDE MDP class, defined at the end of this section.

The second way of dealing with dead ends we consider in this paper is to view them as truly irrecoverable situations and assign $D = \infty$ for visiting them. As a motivation, recall the example of planning a climb to the top of Mount Everest. Since dead ends here cannot be avoided with certainty and the penalty of visiting them is $\infty$, comparing policies based on the expected cost of reaching the goal breaks down — they all have an infinite expected cost. Instead, we would like to find a policy that maximizes the probability of reaching the goal and whose expected cost *over the trajectories that reach the goal* is the smallest.

To describe this policy evaluation criterion more precisely, let $S_t^{\pi_s+}$ be a random variable denoting a distribution over states $s'$ for which $P^\pi(s') > 0$ and in which the MDP may end up if policy $\pi$ is executed starting from state $s$ for $t$ steps. That is, $S_t^{\pi_s+}$ differs from the variable $S_t^{\pi_s}$ used previously by considering *only* states from which $\pi$ can reach the goal. Using the $S_t^{\pi_s+}$ variables, we can mathematically evaluate $\pi$ with two ordered criteria by defining the cost of

---
[1]We implicitly assume that one of the optimal policies is *deterministic Markovian* — a detail we can actually prove but choose to gloss over in this paper for clarity.

a state as an ordered pair

$$J_I^\pi(s) = (P^\pi(s), [J^\pi|P^\pi](s)) \quad (7)$$

$$\text{where}\quad [J^\pi|P^\pi](s) = \mathbb{E}\left[\sum_{t=0}^{\infty} \mathcal{C}(S_t^{\pi_s+}, A_t^{\pi_s})\right] \quad (8)$$

Specifically, we write $\pi(s) \prec \pi'(s)$, meaning $\pi'$ is preferable to $\pi$ at $s$, whenever $J_I^\pi(s) \prec J_I^{\pi'}(s)$, i.e., when either $P^\pi(s) < P^{\pi'}(s)$, or $P^\pi(s) = P^{\pi'}(s)$ and $[J^\pi|P^\pi](s) > [J^{\pi'}|P^{\pi'}](s)$. Notice that the second criterion is used conditionally, only if two policies are equal in terms of the probability of reaching the goal, since maximizing this probability is the foremost priority. Note also that if $P^\pi(s) = P^{\pi'}(s) = 0$, then both $[J^\pi|P^\pi](s)$ and $[J^{\pi'}|P^{\pi'}](s)$ are ill-defined. However, since that means that neither $\pi$ nor $\pi'$ can reach the goal from $s$, we define $[J^\pi|P^\pi](s) = [J^{\pi'}|P^{\pi'}](s) = 0$ for such cases, and hence $J_I^\pi(s) = J_I^{\pi'}(s)$.

As in the finite-penalty case, we can demonstrate that there exists a policy $\pi^*$ that is at least as large as all others at all states under the $\prec$-ordering above, and hence optimal, i.e.

$$\pi^*(s) = \arg\max_{\prec \pi} J_I^\pi(s) \;\forall\; s \in \mathcal{S} \quad (9)$$

We are now ready to capture the above intuitions in a definition of the SSPUDE MDP class and its subclasses fSSPUDE and iSSPUDE:

**Definition** An SSP with Unavoidable Dead Ends (SSPUDE) MDP is a tuple $\langle \mathcal{S}, \mathcal{A}, \mathcal{T}, \mathcal{C}, \mathcal{G}, D, s_0\rangle$, where $\mathcal{S}, \mathcal{A}, \mathcal{T}, \mathcal{C}, \mathcal{G}$, and $s_0$ are as in the SSP MDP definition, $D \in \mathbb{R}^+ \cup \{\infty\}$ is a penalty incurred if a dead-end state is visited. In a SSPUDE MDP, every improper policy must incur an infinite expected cost as defined by Eq. 1 at all states from which it can't reach the goal with probability 1.

If $D < \infty$, the MDP is called an fSSPUDE MDP, and its optimal solution is a policy $\pi^*$ satisfying $\pi^*(s) = \min_\pi J_F^\pi(s)$ for all $s \in \mathcal{S}$.

If $D = \infty$, the MDP is called an iSSPUDE MDP, and its optimal solution is a policy $\pi^*$ satisfying $\pi^*(s) = \max_{\prec \pi} J_I^\pi(s)$ for all $s \in \mathcal{S}$.

Our iSSPUDE class is related to multi-objective MDPs, which model problems with several competing objectives, e.g., total time, monetary cost, etc. [Chatterjee et al., 2006, Wakuta, 1995]. Their solutions are Pareto-sets of all non-dominated policies. Unfortunately, such solutions are impractical due to high computational requirements. Moreover, maximizing the probability of goal achievement converts the problem into a GSSP and hence cannot be easily included in those models. A related criterion has also been studied in robotics [Koenig and Liu, 2002].

## 5 The Case of a Finite Penalty

Equation 5 tells us that for an fSSPUDE instance, the cost of any policy at any state is finite. Intuitively, this implies

that fSSPUDE should be no harder to solve than SSP. This intuition is confirmed by this following result:

**Theorem 3.** *fSSPUDE = SSP.*

*Proof sketch.* To show that every fSSPUDE MDP $M_{fSSPUDE}$ can be converted to an SSP MDP, we augment the action set $\mathcal{A}$ of fSSPUDE with a special action $a'$ that causes a transition to a goal state with probability 1 and that costs $D$. This MDP is an SSP, since reaching the goal with certainty is possible from every state. At the same time, the optimization criteria of fSSPUDE and SSP clearly yield the same set of optimal policies for it.

To demonstrate that every SSP MDP $M_{SSP}$ is also an fSSPUDE MDP, for every $M_{SSP}$ we can construct an equivalent fSSPUDE MDP by setting $D = J^*(s)$. The set of optimal policies of both MDPs will be the same. (Note, however, that the conversion procedure is impractical, since it assumes that we know $J^*(s)$ before solving the MDP.) □

The above conversion from fSSPUDE to SSP immediately suggests solving fSSPUDE with modified versions of standard SSP algorithms, as we describe next.

**Value Iteration:** Theorem 3 implies that $J_F^*$, the optimal cost function of an fSSPUDE MDP, must satisfy the following modified Bellman equation:

$$J(s) = \min\left\{D, \min_{a \in \mathcal{A}}\left[\mathcal{C}(s,a) + \sum_{s' \in \mathcal{S}} \mathcal{T}(s,a,s')J(s')\right]\right\} \quad (10)$$

Moreover, it tells us that $\pi^*$ of an fSSPUDE must be greedy w.r.t. $J_F^*$. Thus, an fSSPUDE can be solved with arbitrarily initialized $VI_{SSP}$ that uses Equation 10 for updates.

**Heuristic Search:** By the same logic as above, all FIND-AND-REVISE algorithms and their guarantees apply to fSSPUDE MDPs if they use Equation 10 in lieu of Bellman backup. Thus, all heuristic search algorithms for SSP work for fSSPUDE.

We note that, although this theoretical result is new, some existing MDP solvers use Equation 10 implicitly to cope with goal-oriented MDPs that have unavoidable dead ends. One example is the miniGPT package [Bonet and Geffner, 2005]; it allows the user to specify a value $D$ and then uses it to implement Equation 10 in several algorithms including $VI_{SSP}$ and LRTDP.

## 6 The Case of an Infinite Penalty

In contrast to fSSPUDE MDPs, no existing algorithm can solve iSSPUDE problems either implicitly or explicitly, so all algorithms for tackling these MDPs that we present in this section are completely novel.

### 6.1 Value Iteration for iSSPUDE MDPs

As for the finite-penalty case, we begin by deriving a Value Iteration-like algorithm for solving iSSPUDE. Finding a policy satisfying Eq. 9 may seem hard, since we are effectively dealing with a multicriterion optimization problem.

Note, however, the optimization criteria are, to a certain degree, independent — we can *first* find the set of policies whose probability of reaching the goal from $s_0$ is optimal, and *then* select from them the policy minimizing the expected cost of goal trajectories. This amounts to finding the optimal goal-probability function $P^*$ first, then computing the optimal cost function $[J^*|P^*]$ conditional on $P^*$, and finally deriving an optimal policy from $[J^*|P^*]$. We consider these subproblems in order.

**Finding $P^*$.** The task of finding, for every state, the highest probability with which the goal can be reached by any policy in a given goal-oriented MDP has been studied before — it is the MAXPROB problem mentioned in the Background section. Solving a goal-oriented MDP according to the MAXPROB criterion means finding $P^*$ that satisfies

$$P^*(s) = 1 \; \forall s \in \mathcal{G} \quad (11)$$
$$P^*(s) = \max_{a \in \mathcal{A}} \sum_{s' \in \mathcal{S}} T(s,a,s')P^*(s') \; \forall s \notin \mathcal{G}$$

As already discussed, this $P^*$ can be found by the $VI_{MP}$ algorithm with an arbitrary initializing heuristic.

**Finding $[J^*|P^*]$.** We could derive optimality equations for calculating $[J^*|P^*]$ from first principles and then develop an algorithm for solving them. However, instead we present a more intuitive approach. Essentially, given $P^*$, we will build a modification $M^{P^*}$ of the original MDP whose solution is exactly the cost function $[J^*|P^*]$. $M^{P^*}$ will have no dead ends, have only actions greedy w.r.t. $P^*$, and have a transition function favoring transitions to states with higher probabilities of successfully reaching the goal. Crucially, $M^{P^*}$ will turn out to be an SSP MDP, so we will be able to find $[J^*|P^*]$ with SSPs' familiar machinery.

To construct $M^{P^*}$, observe that an optimal policy $\pi^*$ for an iSSPUDE MDP, one whose cost function is $[J^*|P^*]$, must necessarily use only actions greedy w.r.t. $P^*$, i.e., those maximizing the right-hand side of Eq. 11. For each state $s$, denote the set of such actions as $\mathcal{A}_s^{P^*}$. We focus on non-dead ends, because for dead ends $[J^*|P^*](s) = 0$, and they will not be part of $M^{P^*}$. By Eq. 11, for each such $s$, each $a^* \in \mathcal{A}_s^{P^*}$ satisfies $P^*(s) = \sum_{s' \in \mathcal{S}} T(s,a^*,s')P^*(s')$. Note that this equality expresses the following relationship between event probabilities:

$$P\left(\begin{matrix}\text{Goal was reached}\\ \text{from } s \text{ via optimal policy}\end{matrix}\right)$$
$$= \sum_{s' \in \mathcal{S}} P\left(\begin{matrix}a^* \text{ caused}\\ s \to s' \text{ transition}\end{matrix} \bigwedge \begin{matrix}\text{Goal was reached}\\ \text{from } s \text{ via optimal policy}\end{matrix}\right),$$

or, in a slightly rewritten form,

$$\sum_{s' \in \mathcal{S}} P\left(\begin{matrix}a^* \text{ caused}\\ s \to s' \text{ transition}\end{matrix} \middle| \begin{matrix}\text{Goal was reached}\\ \text{from } s \text{ via optimal policy}\end{matrix}\right) = 1,$$

where $P\left(\begin{array}{c}a^* \text{ caused}\\s \to s' \text{ transition}\end{array}\middle|\begin{array}{c}\text{Goal was reached}\\\text{from } s \text{ via optimal policy}\end{array}\right) = \frac{T(s,a^*,s')P^*(s')}{P^*(s)}$.

These equations essentially say that if $a^*$ was executed in $s$ and, as a result of following an optimal policy $\pi^*$ the goal was reached, then with probability $\frac{T(s,a^*,s_1)P^*(s_1)}{P^*(s)}$ action $a^*$ must have caused a transition from $s$ to $s_1$, with probability $\frac{T(s,a^*,s_2)P^*(s_2)}{P^*(s)}$ it must have caused a transition to $s_2$, and so on. This means that if we want to find the vector $[J^*|P^*]$ of expected costs of goal-reaching trajectories under $\pi^*$, then it is enough to find the optimal cost function of MDP $M^{P^*} = \langle \mathcal{S}^{P^*}, \mathcal{A}^{P^*}, \mathcal{T}^{P^*}, \mathcal{C}^{P^*}, \mathcal{G}^{P^*}, s_0^{P^*} \rangle$, where $\mathcal{G}^{P^*}$ and $s_0^{P^*}$ (if known) are the same as $\mathcal{G}$ and $s_0$ for the iSSPUDE $M$ that we are trying to solve; $\mathcal{S}^{P^*}$ is the same as $\mathcal{S}$ for $M$ but does not include dead ends, i.e., states $s$ for which $P^*(s) = 0$; $\mathcal{A}^{P^*} = \cup_{s \in \mathcal{S}} \mathcal{A}_s^{P^*}$, i.e., the set of actions consists of all $P^*$-greedy actions in each state; for each $a^* \in \mathcal{A}_s^{P^*}$, $\mathcal{T}^{P^*}(s,a^*,s') = \frac{T(s,a^*,s')P^*(s')}{P^*(s)}$, as above, and $a^*$ is "applicable" only in $s$; and $\mathcal{C}^{P^*}(s,a)$ is the same as $\mathcal{C}$ for $M$, except it is defined only for $a \in \mathcal{A}^{P^*}$.

As it turns out, we already know how to solve MDPs such as $M^{P^*}$:

**Theorem 4.** *For an iSSPUDE MDP $M$ with $P^*(s_0) > 0$, MDP $M^{P^*}$ constructed from $M$ as above is an SSP MDP.*

*Proof sketch.* Indeed, $M^{P^*}$ is "almost" like the original iSSPUDE MDP, but has at least one proper policy because, by construction, it has no dead ends. □

Now, as we know [Bertsekas, 1995], $J^*$ for the SSP $M^{P^*}$ satisfies $J^*(s) = \min_{a \in \mathcal{A}^{P^*}} \mathcal{C}^{P^*}(s,a) + \sum_{s' \in \mathcal{S}} \mathcal{T}^{P^*}(s,a,s')J^*(s')$. Therefore, by plugging in $\frac{T(s,a,s')P^*(s')}{P^*(s)}$ in place of $\mathcal{T}^{P^*}(s,a,s')$ and $[J^*|P^*]$ in place of $J^*$, we can state the following theorem for the original iSSPUDE MDP $M$:

**Theorem 5.** *For an iSSPUDE MDP with the optimal goal-probability function $P^*$, the optimal cost function $[J^*|P^*]$ characterizing the minimum expected cost of trajectories that reach the goal satisfies*

$$[J^*|P^*](s) = 0 \ \forall s \ s.t. \ P^*(s) = 0 \quad (12)$$

$$[J^*|P^*](s) = \min_{a \in \mathcal{A}^{P^*}}\left\{\mathcal{C}(s,a) + \sum_{s' \in \mathcal{S}} \frac{T(s,a,s')P^*(s')}{P^*(s)}[J^*|P^*](s')\right\}$$

**Putting It All Together.** Our construction not only let us derive the optimality equation for $[J^*|P^*]$, but also implies that $[J^*|P^*]$ can be found via VI, as in the case of SSP MDPs [Bertsekas, 1995], over $P^*$-optimal actions and non-dead-end states. Moreover, since the optimal policy for an SSP MDP is greedy w.r.t. the optimal cost function and solving an iSSPUDE MDP ultimately reduces to solving an SSP, the following important result holds:

**Theorem 6.** *For every iSSPUDE MDP, there exists a Markovian deterministic policy $\pi^*$ that can be derived from $P^*$ and $[J^*|P^*]$ for non-dead-end states using*

$$\pi^*(s) = \arg\min_{a \in \mathcal{A}^{P^*}}\left\{\mathcal{C}(s,a) + \sum_{s' \in \mathcal{S}} \frac{T(s,a,s')P^*(s')}{P^*(s)}[J^*|P^*](s')\right\} \quad (13)$$

Combining optimality equations 11 and 12 for $P^*$ and $[J^*|P^*]$ respectively with Equation 13, we present a VI-based algorithm for solving iSSPUDE MDPs, called IVI (Infinite-penalty Value Iteration) in Algorithm 1.

---

**Input:** iSSPUDE MDP $M$
**Output:** Optimal policy $\pi^*$ for non-dead-end states of $M$

1. Find $P^*$ using arbitrarily initialized VI$_{MP}$.

2. Find $[J^*|P^*]$ using arbitrarily initialized VI$_{SSP}$ over $M^{P^*}$ with update equations 12

Return $\pi^*$ derived from $P^*$ and $[J^*|P^*]$ via Equation 13

**Algorithm 1:** IVI

---

### 6.2 Heuristic Search for iSSPUDE MDPs

---

**Input:** iSSPUDE MDP $M$
**Output:** Optimal policy $\pi^*_{s_0}$ for non-dead-end states of $M$ rooted at $s_0$

1. Find $P^*_{s_0}$ using FRET initialized with an admissible heuristic $\hat{P} \geq P^*$

2. Find $[J^*|P^*]_{s_0}$ using FIND-AND-REVISE over $M^{P^*}$ with optimality equations 12, initialized with an admissible heuristic $\hat{J} \leq [J^*|P^*]$.

Return $\pi^*_{s_0}$ derived from $P^*_{s_0}$ and $[J^*|P^*]_{s_0}$ via Equation 13

**Algorithm 2:** SHS

---

As we established, solving an iSSPUDE MDP with VI is a two-stage process, whose first stage solves a MAXPROB MDP and whose second stage solves an SSP MDP. In the Background section we mentioned that both of these kinds of MDPs can be solved with heuristic search; MAXPROB — with the FRET framework, and SSP — with the FIND-AND-REVISE framework. This allows us to construct a heuristic search schema called SHS (Staged Heuristic Search) for iSSPUDE MDPs, presented in Algorithm 2.

There are two major differences between Algorithms 1 and 2. The first one is that SHS produces functions $P^*_{s_0}$ and $[J^*|P^*]_{s_0}$ that are guaranteed to be optimal only over the states visited by some optimal policy $\pi^*_{s_0}$ starting from the initial state $s_0$. Accordingly, the SHS-produced policy $\pi^*_{s_0}$ specifies actions only for these states and does not prescribe any for other states. Second, SHS requires two admissible heuristics to find an optimal (partial) policy, one ($\hat{P}$) being an *upper* bound on $P^*$ and the other ($\hat{J}$) being a *lower* bound on $[J^*|P^*]$.

## 7 Equivalences of Optimization Criteria

The presented algorithms for MDPs with unavoidable dead ends are significantly more complicated than those for MDPs with unavoidable ones. Nonetheless, intuition tells us that for a given tuple $\langle \mathcal{S}, \mathcal{A}, \mathcal{T}, \mathcal{C}, \mathcal{G}, D, s_0 \rangle$, solving it under the infinite-penalty criterion (i.e., as an iSSPUDE) should yield the same policy as solving it under the finite-penalty criterion (i.e., as an fSSPUDE) if in the latter case the penalty $D$ is very large. This can be stated as a theorem:

**Theorem 7.** *For iSSPUDE and fSSPUDE MDPs over the same domain, there exists the smallest finite penalty $D_{thres}$ s.t. for all $D > D_{thres}$ the set of optimal policies of fSSPUDE (with penalty $D$) is identical to the set of optimal policies of iSSPUDE.*

*Proof sketch.* Although the full proof is technical, its main observation is simple — as $D$ increases, it becomes such a large deterrent against hitting a dead end that any policy with a probability of reaching the goal lower than the optimal $P^*$ starts having a higher expected cost of reaching the goal than policies optimal according to iSSPUDE's criterion. □

As a corollary, if we choose $D > D_{thres}$, we can be sure that at any given state $s$, all optimal (J$_\text{F}^*$-greedy) policies of the resulting fSSPUDE will have the same probability of reaching the goal, and this probability is $P^*(s)$ according to the infinite-penalty optimization criterion (and therefore will also have the same conditional expected cost $[J^*|P^*]$)

This prompts a question: what can we say about the probability of reaching the goal of J$_\text{F}^*$-greedy policies if we pick $D \leq D_{thres}$? Unfortunately, in this case different greedy policies may not only be suboptimal in terms of this probability, but even for a fixed $D$ each may have a different, arbitrarily low chance of reaching the goal. For example, consider an MDP with three states, $s_0$ (the initial state), $d$ (a dead end), and $g$ (a goal). Action $a_d$ leads from $s_0$ to $d$ with probability 0.5 and to $g$ with probability 0.5 and costs 1 unit. Action $a_g$ leads from $s_0$ to $g$ also with probability 1, and costs 3 units. Finally, suppose we solve this MDP as an fSSPUDE with $D = 4$. It is easy to see that both policies, $\pi(s_0) = a_d$ and $\pi(s_0) = a_g$, have the same expected cost, 3. However, the former reaches the goal with probability 0.5, while the latter always reaches it. The ultimate reason for this discrepancy is that the policy evaluation criterion of fSSPUDE is oblivious to policy's probability of reaching the goal, and optimizes for this parameter only indirectly, via policy's expected cost.

To summarize, we have two ways of finding an optimal policy in the infinite-penalty case, either by directly solving the corresponding iSSPUDE instance, or by choosing a sufficiently large $D$ and solving the finite-penalty fSSPUDE MDP. We do not know of a principled way to choose $D$, but it is typically easy to guess by inspecting the MDP. Thus, although the latter method gives no a-priori guarantees, it often yields a correct answer in practice.

## 8 Experimental Results

The objective of our experiments was to find out the most practically efficient way of finding the optimal policy in the presence of unavoidable dead ends and infinite penalty for visiting them, by solving an iSSPUDE MDP or an fSSPUDE MDP with a large $D$. To make a fair comparison between these methods, we employ very similar algorithms to handle them. For both classes, the most efficient optimal solution methods are heuristic search techniques, so in our experiments we assume knowledge of the initial state and use only algorithms of this type.

To solve an fSSPUDE, we use the implementation of the LRTDP algorithm, an instance of the FIND-AND-REVISE heuristic search framework for SSPs, available in the miniGPT package [Bonet and Geffner, 2005]. As a source of admissible heuristic state costs/goal-probability values, we choose the maximum of atom-min-forward heuristic [Haslum and Geffner, 2000] and SixthSense [Kolobov et al., 2011]. The sole purpose of the latter is to soundly identify many of the dead ends and assign the value of $D$ to them. (Identifying a state as a dead end may be nontrivial if the state has actions leading to other states.)

Since solving iSSPUDE involves tackling two MDPs, a MAXPROB and an SSP, to instantiate the SHS schema (Algorithm 2) we use two heuristic search algorithms. For the MAXPROB component, we use a specially adapted version [Kolobov et al., 2011] of miniGPT's LRTDP, equipped with SixthSense (note that the atom-min-forward heuristic is cost-based and does not apply to MAXPROB MDPs). For the SSP component, we use miniGTP's LRTDP, as for fSSPUDE, with atom-min-forward; SixthSense is unnecessary because SSP has no dead ends.

Our benchmarks were problems 1 through 6 of the Exploding Blocks World domain from IPPC-2008 [Bryce and Buffet, 2008] and problems 1 through 15 of the Drive domain from IPPC-06 [Buffet and Aberdeen, 2006]. Most problems in both domains have unavoidable dead ends. To set the $D$ penalty for the fSSPUDE model, we examined each problem and tried to come up with an intuitive, easily justifiable value for it. For all problems, $D = 500$ yielded a policy that was optimal under both the finite-penalty and infinite-penalty criterion.

Solving the fSSPUDE with $D = 500$ and iSSPUDE versions of each problem with the above implementations yielded the same qualitative outcome on all benchmarks. In terms of speed, solving fSSPUDE was at least an order of magnitude faster than solving iSSPUDE. The difference in used memory was occasionally smaller, but only because both algorithms visited nearly the entire state space reachable from $s_0$ on some problems. Moreover, in terms of memory as well as speed the difference between solving fSSPUDE and iSSPUDE was the largest (that is, solving iSSPUDE was comparatively the *least* efficient) when the given MDP had $P^*_{s_0}(s) = 1$, i.e. the MDP had no dead ends at all or had only avoidable ones.

Although seemingly surprising, these performance patterns have a fundamental reason. Recall that FRET algorithms, used for solving the MAXPROB part of an iSSPUDE, use the BET operator. BET, for every encountered fixed point $P^\times$ of the Bellman backup operator needs to traverse the transition graph involving all actions greedy w.r.t. $P^\times$, starting from $s_0$. Also, FRET needs to be initialized with an admissible heuristic, in our experiments – SixthSense, which assigns the value of 0 to states it believes to be dead ends and 1 to the rest.

Now, consider how FRET operates on a MAXPROB corresponding to an iSSPUDE instance that does not have any dead ends, i.e. on the kind of iSSPUDE MDPs that, as our experiments show, is most problematic. For such a MAXPROB, there exists only one admissible heuristic function, $\hat{P}(s) = 1$ for all $s$, because $P^*(s) = 1$ for all $s$ and an admissible $\hat{P}$ needs to satisfy $\hat{P}(s) \geq P^*(s)$ everywhere. Thus, the heuristic FRET starts with is actually the optimal goal-probability function, and as a consequence, is a fixed point of the Bellman backup operator. Therefore, to conclude that $\hat{P}$ is optimal, FRET needs to build its greedy transition graph. Observe, however, that since $\hat{P}$ is 1 everywhere, this transition graph includes every state reachable from $s_0$, and uses every action in the MDP! Building and traversing it is very expensive.

The same performance bottleneck, although to a lesser extent, can also be observed on iSSPUDE instances that do have unavoidable dead ends. Building large transition graphs significantly slows down FRET (and hence, SHS) even when $P^*$ is far from being 1 everywhere.

The above reasoning may explain why solving iSSPUDE is slow, but by itself does not explain why solving fSSPUDE is fast in comparison. For instance, we might expect the performance of FIND-AND-REVISE algorithms on fSSPUDE to suffer in the following situations. Suppose state $s$ is a dead end not avoidable from $s_0$ by any policy. This means that $J^*(s) = D$ under the finite-penalty optimization criterion, and that $s$ is reachable from $s_0$ by any optimal policy. Thus, FIND-AND-REVISE will halt no earlier than the cost of $s$ under the current cost function reaches $D$. Moreover, suppose that the heuristic $\hat{J}$ initializes the cost of $s$ to 0 — this is one of the possible admissible costs for $s$. Finally, assume that all actions in $s$ lead back to $s$ with probability 1 and cost 1 unit. In such a situation, an FIND-AND-REVISE algorithm will need to update the cost of $s$ $D$ times before convergence. Clearly, this will make the performance of FIND-AND-REVISE very bad if the chosen value of $D$ is very large. This raises the question: was solving fSSPUDE in the above experiments so much more efficient than solving iSSPUDE due to our choice of (a rather small) value for $D$?

To dispel these concerns, we solved fSSPUDE instances of the aforementioned benchmarks with $D = 5 \cdot 10^8$ instead of 500. On all of the 21 problems, the increase in speed compared to the case of fSSPUDE with $D = 500$ was no more than a factor of 1.5. The reason for such a small discrepancy is the fact that, at least on our benchmarks, FIND-AND-REVISE almost never runs into the pathological case described above thanks to the atom-min-forward and SixthSense heuristics. They identify the majority of dead ends encountered by LRTDP and immediately set their costs to $D$. Thus, instead of spending many updates on such states, LRTDP gets their optimal costs in just one step. To test this explanation, we disabled these heuristics and assigned the cost of 0 to all states at initialization. As predicted, the solution time of the fSSPUDE instances skyrocketed by orders of magnitude.

The presented results appear to imply an unsatisfying fact — on iSSPUDE MDPs that are SSPs, the presented algorithms for solving iSSPUDE are not nearly as efficient as algorithmic schema for SSPs, such as FIND-AND-REVISE. The caveat, however, is that the price SSP algorithms pay for efficiency is *assuming* the existence of proper solutions. iSSPUDE algorithms, on the other hand, implicitly *prove* the existence of such a solution, and are therefore theoretically more robust.

# 9 Conclusion

A significant limitation of SSP MDPs is their inability to model dead-end states, consequences of catastrophic action outcomes that make reaching the goal impossible. While attempts to incorporate dead ends into SSP have been made before, a principled theory of goal-oriented MDPs with dead-end states has been lacking.

In this paper, we present new general MDP classes that subsume SSP and make increasingly weaker assumptions about the presence of dead ends. SSPADE assumes that dead ends are present but an agent can avoid them if it acts optimally from the initial state. fSSPUDE admits unavoidable dead ends but expects that an agent can put a finite price on running into a dead end. iSSPUDE MDPs model scenarios in which entering a dead end carries an infinite penalty and is to be avoided at all costs.

For these MDP classes, we present VI-based and heuristic search algorithms. We also study the conditions under which they have equivalent solutions. Our empirical results show that, in practice, solving fSSPUDE is much more efficient and yields the same optimal policies as iSSPUDE.

In the future, we hope to answer the question: are iSSPUDE MDPs fundamentally harder than fSSPUDE, or can we invent more efficient heuristic search algorithms for them? Besides, as we found out after submitting this article, a more general class of MDPs than iSSPUDE, called $S^3P$, has been proposed in the literature but not completely solved [Teichteil-Königsbuch, 2012]. Deriving algorithms for it will be our next research objective.

**Acknowledgments.** This work has been supported by NSF grant IIS-1016465, ONR grant N00014-12-1-0211, and the UW WRF/TJ Cable Professorship.